\pdfoutput=1

\documentclass[11pt]{article}

\usepackage{acl2023}

\usepackage{times}
\usepackage{latexsym}

\usepackage[T1]{fontenc}

\usepackage[utf8]{inputenc}

\usepackage{microtype}

\usepackage{inconsolata}

\usepackage{amsmath,amssymb,amsfonts}
\usepackage{algorithmic}
\usepackage{graphicx}
\usepackage{textcomp}
\usepackage{xcolor}
\usepackage{listings}
\usepackage{comment}
\usepackage{caption}
\usepackage{subcaption}
\usepackage{lipsum}
\newcommand{\footURL}[1]{\footnote{\url{#1}}}
\usepackage[utf8]{inputenc}
\usepackage{threeparttable}
\usepackage{multirow}
\usepackage{hhline}
\usepackage{makecell}
\usepackage{tabularx}
\newcolumntype{Y}{>{\centering\arraybackslash}X}
\newcolumntype{Z}{>{\raggedleft\arraybackslash}X}

\def\BibTeX{{\rm B\kern-.05em{\sc i\kern-.025em b}\kern-.08em
    T\kern-.1667em\lower.7ex\hbox{E}\kern-.125emX}}
\usepackage{hyperref} 
\hypersetup{
colorlinks   = true,
citecolor    = blue,
urlcolor    =  blue
}
\title{Sinhala-English Word Embedding Alignment: Introducing Datasets and Benchmark for a Low Resource Language}

\author{Kasun Wickramasinghe \and Nisansa de Silva \\
 Department of Computer Science \& Engineering \\ University of Moratuwa, Katubedda 10400, Sri Lanka\\
        \texttt{\{kasunw.22,NisansaDdS\}@cse.mrt.ac.lk}}

\begin{document}
\maketitle
\begin{abstract}
Since their inception, embeddings have become a primary ingredient in many flavours of Natural Language Processing (NLP) tasks supplanting earlier types of representation. Even though multilingual embeddings have been used for the increasing number of multilingual tasks, due to the scarcity of parallel training data, low-resource languages such as Sinhala, tend to focus more on monolingual embeddings. Then when it comes to the aforementioned multilingual tasks, it is challenging to utilize these monolingual embeddings given that even if the embedding spaces have a similar geometric arrangement due to an identical training process, the embeddings of the languages considered are not aligned. This is solved by the embedding alignment task. Even in this, high-resource language pairs are in the limelight while low-resource languages such as Sinhala which is in dire need of help seem to have fallen by the wayside. In this paper, we try to align Sinhala and English word embedding spaces based on available alignment techniques and introduce a benchmark for Sinhala language embedding alignment. In addition to that, to facilitate the supervised alignment, as an intermediate task, we also introduce Sinhala-English alignment datasets. These datasets serve as our anchor datasets for supervised word embedding alignment. Even though we do not obtain results comparable to the high-resource languages such as French, German, or Chinese, we believe our work lays the groundwork for more specialized alignment between English and Sinhala embeddings.
\end{abstract}

\section{Introduction} \label{introduction}

Embedding spaces have been shown to have similar geometric arrangements~\cite{mikolov2013exploiting, conneau2017word} especially when the training process is similar but, separately trained spaces are not aligned by default and that is a huge burden when it comes to certain multilingual tasks where having aligned embeddings are required.

Aligned embeddings are useful in multilingual tasks since similar words and sentences in each language can be considered to reside closer to each other in a common embedding space. So that we can do mathematical operations on the embeddings regardless of the language~\cite{feng-etal-2022-language, lample2019cross}. 

The alignment is required for two types of embedding models: 
\begin{enumerate}
    \item Embedding models separately trained on monolingual data~\cite{mikolov2013efficient, bojanowski-etal-2017-enriching} and
    \item Multilingual embedding models trained on parallel multilingual data~\cite{feng-etal-2022-language, lample2019cross, conneau-etal-2020-unsupervised}.
\end{enumerate}  

As far as the multilingual models are concerned, most of the time the training process itself implicitly encourages alignment~\cite{feng-etal-2022-language, lample2019cross}. Conversely, when the monolingual models are concerned, the alignment has to be done explicitly after the models are trained. Multilingual models~\cite{feng-etal-2022-language, lample2019cross,conneau-etal-2020-unsupervised} are becoming more common for multilingual tasks nowadays due to the aforementioned implicit alignment of the training process~\cite{feng-etal-2022-language, lample2019cross}.

Monolingual embedding models have been there for decades and aligning monolingual embedding models is beneficial in various aspects rather than using multilingual models.
\begin{itemize}
    \item Monolingual models are lightweight
    \item Can be run using simpler libraries and frameworks
    \item Using multilingual models may be redundant due to supporting many languages~\cite{feng-etal-2022-language, lample2019cross, conneau-etal-2020-unsupervised}
    \item Multilingual model accuracy can be compromised due to the support of many languages~\cite{feng-etal-2022-language}
    \item The accuracy for low-resource languages can be less compared to high-resource languages due to training data imbalance~\cite{feng-etal-2022-language} in multilingual models (Eg: $\sim$700 Sinhala tokens in XLM-R~\cite{conneau-etal-2020-unsupervised} vocabulary)
    \item Training or fine-tuning a multilingual model is time and resource-consuming~\cite {feng-etal-2022-language, lample2019cross, conneau-etal-2020-unsupervised}
\end{itemize}

Therefore, aligning existing monolingual models is still vital. \emph{Aligned} word embedding models for common high-resource languages are officially provided by FastText\footURL{https://bit.ly/3LGqDrE} but most of the aligned low-resource language models are not publicly available. 
\emph{Sinhala} being such a low-resource language, suffers from the aforementioned difficulties~\cite{de2019survey,ranathunga-de-silva-2022-languages}. Several related works to the Sinhala language have been done previously by~\citet{smith2016offline} using \textit{Procrusts} and~\citet{9525667} using \textit{VecMap} but, our attempt to properly make everything ready and available for future research. Therefore, our effort here is to,
\begin{itemize}
    \item Set a benchmark for Sinhala word embedding alignment
    \item Introduce dataset induction methods for low-resource languages when parallel word corpora are not available
    \item Introduce MUSE\footURL{https://github.com/facebookresearch/MUSE}-like~\cite{conneau2017word} alignments datasets for Sinhala-English language pair
    \item Provide aligned embeddings for Sinhala-English pair
    \item Release the code-base\footURL{https://bit.ly/3t3SKu7} related to all the experiments we have conducted.
\end{itemize}

This is more so the case for low-resource languages such as Sinhala~\cite{de2019survey}. This problem gets further accentuated due to the unreliable nature of the quality of existing parallel corpora for such low-resource languages~\cite{kreutzer-etal-2022-quality}. 

\section{Related Work}

\subsection{Embedding Generation}


The first major turning point in the word embedding domain was the introduction of Word2Vec by \citet{mikolov2013efficient}.
%
%
Subsequently, two new Word2Vec-like embedding models were released which are the well-known Glove~\cite{pennington-etal-2014-glove} and FastText~\cite{bojanowski-etal-2017-enriching} models. Those are global embedding models.

The idea behind \emph{Embeddings from Language Models (ELMo)}~\cite{peters-etal-2018-deep} is generating a \emph{context-based embedding} for a given word.
%
In the transformers~\citet{vaswani2017attention} era, the first member of context-based transformer encoders is the \emph{BERT}~\cite{devlin-etal-2019-bert} which is a stack of transformer encoders trained on two objectives named \emph{Masked Language Modeling (MLM)} and \emph{Next Sentence Prediction}. After that many variants of \emph{BERT} have been released including \emph{sentence transformers}~\cite{reimers-gurevych-2019-sentence}.

\subsection{Word Embedding Alignment Techniques} \label{vailable_alignment_techniques}
For word embedding alignment, there have been different approaches since the release of Word2Vec~\cite{mikolov2013efficient}. The first work we come across is the work by~\citet{mikolov2013exploiting} in 2013. In the following subsections, we are talking about the major approaches that have been there for word embedding alignment.

\subsubsection{Simple Linear Mapping} \label{method_l2}
Our first method is to find a linear mapping $W$, assuming the geometric arrangements of two embedding spaces are similar as per \citet{mikolov2013exploiting}. The optimizing objective, therefore, is to minimize the Euclidean distance between the target and the mapped vectors as per Equation~\ref{eq1}.
\begin{equation}
\underset{W}{min}\sum_{i=1}^{n} \left \| Wx_{i} -z_{i} \right \|^{2}\label{eq1}
\end{equation}

\subsubsection{Orthogonal Mapping} \label{method_ortho}
The second method we are trying is, finding an \emph{orthogonal} mapping between the \emph{normalized} source and the target embedding spaces \cite{xing-etal-2015-normalized}. 
The major improvement we can expect from this mapping is that the optimizing objective is, from one perspective, optimizing the cosine distance between the target and the mapped embedding. The optimizing objective is as per Equation~\ref{eq2}.
\begin{equation}
\underset{W}{max}\sum_{i}(Wx_{i})^{T}z_{i} \label{eq2}
\end{equation}

\subsubsection{Orthogonal Procrustes Mapping} \label{method_proc}
In this case, the orthogonal transformation matrix is approximated using the product $UV^{T}$, where $U$ and $V$ are the transformation matrices of singular value decomposition (SVD) of the product $X^{T}Y$ where $X$ and $Y$ are the original source and target embeddings \cite{smith2016offline}. As we know the $U$ and $V^{T}$ matrices only perform translation, rotation, uniform scaling, or a combination of these transformations, and no deformations are performed. Therefore the $UV^{T}$ will simply align one embedding space to the other with the assumption that the geometric arrangement of the two spaces is similar.

\subsubsection{CSLS Optimization} \label{method_csls}
The third method we are trying is minimizing the \emph{Cross-domain similarity local scaling} (CSLS) loss (Equation~\ref{eq3}) as the optimization criterion \cite{joulin-etal-2018-loss}. The mapping is assumed to be \emph{orthogonal} and the emending is assumed to be \emph{normalized}.

\begin{equation}\label{eq3}
\begin{aligned}
\underset{W\in O_{d}}{min} \frac{1}{n} \sum_{i=1}^{n}-2x_{i}^{T}W^{T}y_{i} + \\
\frac{1}{k}\sum_{y_{j}\in \mathit{N_{Y}}(Wx_{i})}x_{i}^{T}W^{T}y_{j} + \\
\frac{1}{k}\sum_{Wx_{j}\in \mathit{N_{X}}(y_{i})}x_{j}^{T}W^{T}y_{i}
\end{aligned}
\end{equation}

\citet{joulin-etal-2018-loss} have addressed the so-called \emph{hubness problem} in embedding alignment. \emph{Hubs} are words that appear too frequently in the neighbourhoods of other words. There have been solutions to mitigate this issue at \emph{inference} by using different criteria (loss) such as Inverted Softmax (IFS) or CSLS, rather than using the same criteria used at the training phase. Using different criteria for inference adds an inconsistency. Therefore \citet{joulin-etal-2018-loss} have included the CSLS criteria directly to the training objective and have achieved better results compared to previous related work. This is one of the alignment techniques used by \emph{FastText} for their official \emph{aligned word vectors}.

\subsubsection{Unsupervised Techniques} \label{method_unsup}
The fourth method we are trying is the unsupervised alignment method where a parallel dictionary is not needed for the alignment where creating a quality parallel dictionary may consume extra time and resources. Unsupervised alignment can be done using,
\begin{itemize}
    \item \textbf{Traditional statistical optimization techniques:}
    \citet{artetxe-etal-2018-robust} use an unsupervised initialization for the seed words based on the word similarity distributions claiming that the similar words of two languages should have similar distributions and then improve the mapping in an iterative manner using a self-learning technique. This method has been published as a framework called \emph{VecMap}\footURL{https://github.com/artetxem/vecmap}.
    
    The work by \citet{grave2019unsupervised} is about Procrustes analysis which learns a linear transformation between two sets of matched points $X \in R^{nXd}$ and $Y \in R^{nXd}$. If the correspondences between the two sets are known (i.e., which point of $X$ corresponds to which point of $Y$), then the linear transformation can be recovered using least square minimization or finding the orthogonal mapping between the two spaces just like in supervised methods described just above.
    In this case, we do not know the correspondence between the two sets, nor the linear transformation. Therefore, the goal is to learn an orthogonal matrix $Q \in O_{d}$, such that the set of points $X$ is close to the set of points $Y$ and 1-to-1 correspondences (permutation matrix) can be found. They use the \emph{Wasserstein distance} or \emph{Earth Mover Distance}  as the measure of distance between our two sets of points and then combine it with the orthogonal Procrustes, leading to the problem of Procrustes in Wasserstein distance or Wasserstein Procrustes (WP).

    \citet{aboagye-etal-2022-quantized} have proposed Quantized Wasserstein Procrustes (qWP) Alignment which reduces the computational cost of the permutation matrix approximation in WP by quantizing the source and target embedding spaces.
    
    \item \textbf{Adversarial methods:} One of the well-known unsupervised techniques is adversarial techniques where a \emph{Generator} tries to mimic the desired results while a \emph{Discriminator} tries to distinguish the real results from the generator results. The contest between the \emph{Generator} and the \emph{Discriminator} ends up having a \emph{Generator} that can generate almost similar real results which the \emph{Discriminator} can no longer distinguish. The work by \citet{conneau2017word} follows an adversarial approach where they have obtained similar accuracy numbers as supervised alignment techniques by then.
\end{itemize}

\subsection{English-Sinhala Embedding Alignment}
\citet{smith2016offline} have published\footURL{https://bit.ly/3PTRW3Y} EN-Si alignment matrix along with 77 other languages. However, they have only worked in the Si$\rightarrow$En direction (i.e. mapping En as the target). Their alignment datasets have not been published and most of the later experiments have been done using the \emph{MUSE} datasets. Both MUSE and \citet{smith2016offline} not having published an En-Si dataset we have to create our own dataset for supervised alignments as well as alignment result evaluation. Recently \citet{9525667} have experimented VecMap to align English and Sinhala embedding spaces for lexicon induction task

\subsection{Alignment Datasets} \label{lit-datasets}
The works by~\citet{guzman-etal-2019-flores},~\citet{hameed2016automatic}, \citet{banon-etal-2020-paracrawl} and~\citet{vasantharajan2022adapting} that are comprised of sentence and paragraph level parallel entries. Apart from that there are several sentence and document-level parallel corpora available in OPUS\footURL{https://opus.nlpl.eu/}. They are well suited for higher-level multilingual tasks like Machine Translation (MT) but, not for lower-level tasks like word embedding alignment.

When it comes to word-level parallel corpora or simply dictionaries, we can find very few open-source resources for English-Sinhala language pairing. For most of the common language pairs, common alignment datasets have been published by MUSE but Sinhala is not available there. The dictionary \emph{Subasa Ingiya}\footURL{https://subasa.lk/?page_id=3738}~\cite{wasala2008ensitip} is one of them which is a small dictionary that contains about 36000 pairs and contains not only word pairs but also phrases. The next resource is by \citet{wickramasinghe2023sinhala} which introduces several pure word-level dictionaries.

\section{Methodology}
In this section, we present the methodologies we followed to obtain,
\begin{enumerate}
    \item An alignment dataset for supervised embedding alignment
    \item The alignment matrix between English and Sinhala word embedding spaces
\end{enumerate}

%
Our primary research objective is to have an aligned Sinhala word embedding space with another high-resource language word embedding space such as English. We are experimenting with some of the techniques mentioned in Section~\ref{vailable_alignment_techniques}. For the supervised techniques we need a parallel word corpus where each parallel pair acts as so-called \emph{Anchor words}. For that purpose, we are creating an English-Sinhala parallel word dictionary which is our first task. The results we obtained and comparison with existing results are presented in Section~\ref{experiments}.

\subsection{Alignment Dataset Creation} \label{dataset_creation}
Our first task is to create an alignment dataset for the supervised alignment. We experimented with two statistical methods and one available dataset adaptation to form the parallel word dictionary alias, our \emph{alignment dataset}. In this section, we are presenting those techniques.

\subsubsection{Pointwise Mutual Information Criterion} \label{dict_pmi}

Pointwise Mutual Information (PMI) is used to identify how given two events are associated with each other. In Natural Language Processing (NLP) this measure is slightly improved as positive PMI where negative PMI values are clipped to 0 and this measure is used to identify context words of a given word.

\begin{equation} \label{pmi_equation}
\begin{split}
pmi(x, y) &= log_{2}\left(\frac{P(x, y)}{P(x)P(y)}\right) \\
&= log_{2}\left(\frac{N.count(x, y)}{count(x).count(y)}\right)
\end{split}
\end{equation}

\begin{equation}
\begin{aligned}
&ppmi(src, tgt) = max\left\{pmi(src, tgt), 0 \right\} \\
&= max\left\{log_{2}\left(\frac{N.count(src, tgt)}{count(src).count(tgt)}\right), 0  \right\}  \label{ppmi_equation}
\end{aligned}
\end{equation}

We used the PPMI measure between source and target word pairs in several parallel English-Sinhala corpora and by applying a threshold to PPMI we tried to obtain the corresponding translation (i.e. target word) for each source word.

Even if there are many sentence and paragraph-level parallel corpora out there, by considering the \emph{size} and \emph{quality (alignment)}, we selected only the following English-Sinhala parallel corpora to extract the dictionaries.
\begin{enumerate}
    \item CCAligned-v1\footURL{https://bit.ly/3PECd7P} - by~\citet{el-kishky-etal-2020-ccaligned}
    \item OpenSubtitles-v2018\footURL{https://bit.ly/3PEE5xh} - Initially by~\citet{tiedemann-2016-finding} 
\end{enumerate}

In our case, the $N$ should be the total number of data points in the parallel corpus. Hence it becomes a global context rather than a local context. We observed that the dictionary building becomes unstable, i.e. many false pairs along with few correct pairs in the result. Therefore, we experimented with another approach that pays more attention to the local context. 

\subsubsection{Conditional Probability Product} \label{dict_prob}

In this approach, we have made a simple but valid assumption. That is, \emph{"In a parallel corpus, the corresponding word translation pairs should co-occur"}. In other words, \emph{"If two source and target language words co-occur more often, then there is a high chance for them to be a translation pair"}. If we can have a large enough corpus then we can say that this measurement tends to be more accurate due to the sampling statistics being closer to population statistics. Based on this assumption, we can find word translation pairs,
as utilized in the corresponding optimization criterion  in Equation~\ref{conditioal_max_criteria}, by finding the source-target word pairs that maximize the product of the two conditional probabilities:
\begin{enumerate}
    \item Finding the target word in the context of the source word (corresponding translation) given the source word - $P(target|source)$ 
    \item Finding the source word in the context of the target word (corresponding translation) given the target word - $P(source|target)$ 
\end{enumerate}

\begin{equation} \label{conditioal_max_criteria}
\begin{split}
&\underset{src, tgt}{max} \left[P\left(src|tgt\right)P\left(tgt|src\right)\right] \\
&\Longrightarrow \underset{src, tgt}{max} \left[\frac{P(src,tgt)^{2}}{P(source)P(target)}\right] \\
&\Longrightarrow \underset{src, tgt}{max} \left[\frac{count(src,tgt)^{2}}{count(src).count(tgt)}\right]
\end{split}
\end{equation}

We used the same two corpora, \emph{CCAligned} and \emph{OpenSubtitles}, used in \emph{ppmi} method explained in Section~\ref{dict_pmi} to build the dictionaries here as well. This dataset is referred to as \emph{Prob-based-dict} throughout the paper.

\subsubsection{Using an Available Dataset} \label{dict_avaibale}

Recent work by \citet{wickramasinghe2023sinhala} has introduced three English-Sinhala parallel dictionary datasets and the FastText version of that can be used for our work directly. They have published the datasets in GitHub\footURL{https://bit.ly/3ZDPrX2}.

Subsets of their dataset have been used to perform the embedding alignment. When building the alignment dataset we used 5k unique source words in the trainset and 1.5k unique source words in the test set. Not only that in the training set, we built the dataset purposefully including the most frequent English and Sinhala words. That is how MUSE datasets have been built as well. 
The datasets derived from this have been referred to with \emph{En-Si-para} and \emph{Si-En-para} prefixes in the paper.

\subsubsection{Dataset Statistics}
The statistics of the dataset are shown in Table~\ref{stat_table}. We have shown the unique word percentage with and without stop-words and, the lookup-precision with respect to the FastText~\cite{bojanowski-etal-2017-enriching,joulin-etal-2017-bag} vocabularies as described in Equation~\ref{lookup_precision_eqn}. Spacy\footURL{https://spacy.io/models/en}(En) and work by~\citet{lakmal-etal-2020-word} (Si) have been used for stop-word removal wherever necessary. 

The \emph{Look-up Precision}, $P_{L}$ means, the proportion of \emph{a word present in the FastText vocabulary, given that word is present in our alignment dictionary}. It is explained in Equation~\ref{lookup_precision_eqn}. The same thing can be simplified according to Equation~\ref{coverage_eqn} where $N_{vocab}$ is the alignment dataset vocabulary size and $N_{available}$ is the number of dataset vocabulary words available in FastText vocabulary.

\begin{equation} 
\label{lookup_precision_eqn}
\begin{aligned}
P_{L} = P\Big(\frac{\text{word present in the FastText vocabulary}}{\text{word present in the dictionary}}\Big)
\end{aligned}
\end{equation}

\begin{equation} \label{coverage_eqn}
\begin{aligned}
P_{L} = coverage = \frac{N_{available}}{N_{vocab}}
\end{aligned}
\end{equation}


\begin{table*}[!htb]
  \centering
  
      \begin{tabularx}{\textwidth}{ll|ZZ|ZZ|ZZ|ZZ}
        \hline
        \multirow{2}{*}{Dataset} & \multirow{2}{*}{Language} & \multicolumn{2}{c}{Entries} & \multicolumn{2}{c}{\makecell{Unique\% \\ w.r.t. stopwords}}  & \multicolumn{2}{c}{\makecell{$P_{L}$\% \\(Each language)}} & \multicolumn{2}{c}{\makecell{$P_{L}$\% \\(Both languages)}}\\
        \hhline{~~--------}
         &  & Unique  & Total  & With  & Without  &  wiki\textsuperscript{*} & cc\textsuperscript{\textdaggerdbl} & wiki\textsuperscript{*} & cc\textsuperscript{\textdaggerdbl} \\
        
    
        \hline
        \multirow{2}{5em}{Prob-based-dict} & English & 36713 & 67404 & 54.47 & 54.78 & 99.99 & 99.99 & \multirow{2}{3em}{47.70} &  \multirow{2}{3em}{99.99}\\
        & Sinhala & 53612 & 67404 & 79.54 & 79.67 & 39.22 & 100 &  & \\
        
    
        
        \hline
        \multirow{2}{5em}{en-si-para-cc-5k\textsuperscript{\textdagger}} & English & 5000 & 12803 & 39.05 & 39.67 & 100.00 & 100.00 & \multirow{2}{3em}{100.00} & \multirow{2}{3em}{100.00}\\
        & Sinhala & 11403 & 12803 & 89.07 & 89.21 & 100.00 & 100.00 &  & \\
    
        \hline
        \multirow{2}{5em}{en-si-para-wiki-5k\textsuperscript{\textdagger}} & English & 5000 & 12782 & 39.12 & 39.74 & 100.00 & 100.00 & \multirow{2}{3em}{100.00} & \multirow{2}{3em}{100.00}\\
        & Sinhala & 11394 & 12782 & 89.14 & 89.28 & 100.00 & 100.00 &  & \\
        \hline

        \multirow{2}{5em}{si-en-para-cc-5k\textsuperscript{\textdagger}} & English & 2406 & 6113 & 39.36 & 40.73 & 100.00 & 100.00 & \multirow{2}{3em}{100.00} & \multirow{2}{3em}{100.00}\\
        & Sinhala & 5000 & 6113 & 81.81 & 81.96 & 100.00 & 100.00 &  & \\
    
        \hline
        \multirow{2}{5em}{si-en-para-wiki-5k\textsuperscript{\textdagger}} & English & 2397 & 6104 & 39.27 & 40.63 & 100.00 & 100.00 & \multirow{2}{3em}{100.00} & \multirow{2}{3em}{100.00}\\
        & Sinhala & 5000 & 6104 & 81.93 & 82.09 & 100.00 & 100.00 &  & \\
        \hline

      \end{tabularx}

      \caption{Dataset Statistics: Statistics of the alignment datasets we have experimented with\newline
      \textsuperscript{*} w.r.t. wiki-based FastText vocabulary
    \textsuperscript{\textdaggerdbl} w.r.t. common-cawl FastText vocabulary\\
    \textsuperscript{\textdagger} Subsets of \citet{wickramasinghe2023sinhala} }
    \label{stat_table}
\end{table*}

\subsection{Embedding Alignment}
We have conducted the embedding alignment with FastText embeddings for English (En) (cc\footURL{https://bit.ly/3PCZ6It}, wiki\footURL{https://bit.ly/46aJzGX}) and Sinhala (Si) (cc\footURL{https://bit.ly/2JXAyL8}, wiki\footURL{https://bit.ly/48BGFgf}) trained on \emph{Common Crawl}\footURL{https://commoncrawl.org/} (cc) and \emph{Wikipedia}\footURL{https://www.wikipedia.org/} (wiki) with the same setups followed by \citet{joulin-etal-2018-loss}.
\begin{itemize}
    \item Learning rate in \{1, 10, 25, 50\} and number of epochs in \{10, 20\}
    \item Center the word vectors (optional)
    \item The number of nearest neighbours in the CSLS loss is 10
    \item Use the l2-normalized word vectors
    \item Use 200k word vectors for the training
\end{itemize}
We adopted our scripts from the alignment scripts by MUSE and  FastText\footURL{https://bit.ly/3Zz21Xe}.
One major observation was that when we use an alignment dataset that consists of the most common words in languages, we obtain a higher test accuracy than having an alignment dataset without considering the most frequent words in languages.

\section{Experiments} \label{experiments}

In this section, we present the experiments we have conducted and the obtained results and observations. We are using the FastText official embeddings of \cite{bojanowski-etal-2017-enriching}. FastText provides two main embedding models: 1) Embeddings trained on Wikipedia (\textit{wiki}), 2) Embeddings trained on Common-Crawl (\textit{cc}).

Most of the previous related work has been done using the wiki embeddings but, when it comes to Sinhala wiki FastText embeddings, there are only 79030 word vectors in the official model (this is because the Sinhala content on Wikipedia is very low: To get an idea, the number of English articles at the moment are more than 6.5M while the number of Sinhala articles are just around 20k) but, the cc Sinhala model contains 808044 word vectors and therefore the wiki vectors are not rich enough for Sinhala. The experimental results also prove that fact. Due to that fact, in some comparisons, we are presenting the results obtained from the cc model.

Sinhala is morphologically richer than English and therefore the alignment is comparatively difficult. In most cases, a single English word can have multiple Sinhala representations. In that case, it is \emph{not a good measure} to check the @1 precision on the test set to evaluate the alignment quality. Therefore checking a higher top-k precision (like @5 or @10) will be a better measure. The Procrustes alignment 
evaluation by \citet{smith2016offline} also shows comparatively low @1 precision for Sinhala (language code Si - recall that they have performed the Si$\rightarrow$En mapping).
According to \citet{aboagye-etal-2022-quantized} results, work by  \citet{joulin-etal-2018-loss} gives the best alignment results and therefore we have used \citet{joulin-etal-2018-loss} as the main reference paper for our work here.

\subsection{Dataset Comparison} \label{exp_dataset_comparison}
As explained in section~\ref{dataset_creation}, we have created the alignment datasets in 3 different approaches, PPM based, conditional probability-based, and using a subset of the dataset by~\citet{wickramasinghe2023sinhala}. In the first experiment, we evaluated all the datasets by aligning the English and Sinhala embeddings using the Procrustes (see section~\ref{method_proc}) method. The results are shown in Table~\ref{alignment_results_datasets}. 

\begin{table}[!ht]
    \centering
   
    \begin{tabular}{l|ll}%
    \hline
        \multirow{2}{*}{Dataset} &\multicolumn{2}{c}{Retrieval} \\
        \hhline{~--}
        & NN & CSLS \\
        \hline
        Prob-based-dict & 13.6 & 16.7 \\ \hline
        En-Si-para-cc-5k & \textbf{16.4} & \textbf{20.4} \\ \hline
    \end{tabular}
     \caption{En-Si Procrustes Embedding Alignment Results of cc-Fasttext embeddings on different datasets}
    \label{alignment_results_datasets}
\end{table}

We can see that the best accuracies have been shown by the \emph{En-Si-para-cc-5k} and \emph{En-Si-para-wiki-5k} datasets and therefore, for the rest of the experiments we have used the datasets created using \citet{wickramasinghe2023sinhala} dataset. 

\subsection{Alignment Results} \label{alignment_results}
\begin{figure}[hbt!]
    
     \centering
     \begin{subfigure}{0.5\textwidth}
         
         \centering
         \centerline{\includegraphics[width=\textwidth]{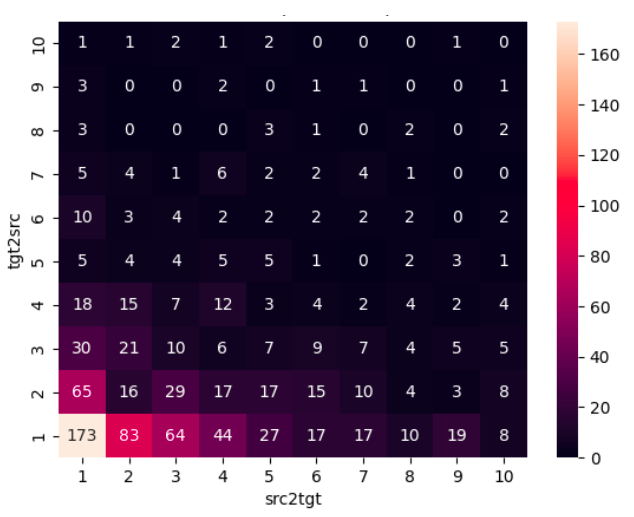}}
         \caption{Retrieval distribution with NN criteria}
         \label{headmap_nn}
     \end{subfigure}
     \hfill
     \begin{subfigure}{0.5\textwidth}
         
         \centering
         \centerline{\includegraphics[width=\textwidth]{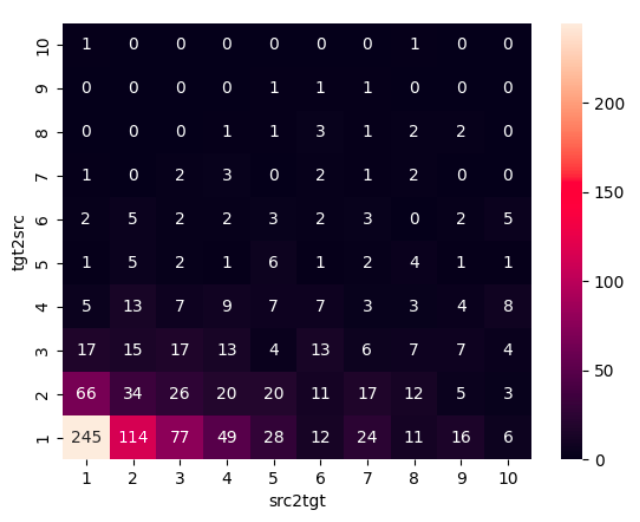}}
         \caption{Retrieval distribution with CSLS criteria}
         \label{headmap_csls}
     \end{subfigure}

     \caption{Top-k Retrieval distribution for RCSL alignment. (The numbers indicate how many pairs in the test set are retrieved in En$\rightarrow$Si and Si$\rightarrow$En directions with corresponding top-k values)}
    \label{evaluation_heatmap}
\end{figure}

Table~\ref{precision_results_wiki_cc} reports the look-up/translation precision of the aligned wiki and cc English-Sinhala embeddings with different alignment techniques and retrieval criteria. The term after the last plus sign is the retrieval criteria. We can see that cc vectors show better alignment than wiki vectors.
Table~\ref{alignment_methods_comparison} shows the translation precision of different alignment techniques. RCSLS gives the best alignment in En$\rightarrow$Si direction while the refined Procrustes method gives the best accuracy in Si$\rightarrow$En direction.
Table~\ref{alignment_results_other_si_work} shows a comparison between the Si-En alignment performed by \citet{smith2016offline}. They have reported the alignment results in Si$\rightarrow$En direction only and also provided the alignment matrix associated with the alignment. The evaluation done using that alignment matrix and our evaluation dataset (rows 2, 3 of Table~\ref{alignment_results_other_si_work}) may not reflect the exact accuracy since the original alignment dataset used by \citet{smith2016offline} is not published and, therefore we cannot guarantee that our evaluation set and their training set are disjoint.
Table~\ref{retrieval_methods_comparison} in Appendix~\ref{appendix-retrieval-methods} has further relevant analysis.
Figure~\ref{evaluation_heatmap} shows the top-k retrieval distribution in both source-target and target-source directions of the aligned embeddings on the test sets for RCSL+NN and RCSL+CSLS using cc-FastText embeddings.

\begin{table*}[!ht]
    \centering
    \resizebox{2\columnwidth}{!}{%
    \begin{tabular}{p{4cm}|p{0.6cm}p{0.6cm}p{0.75cm}|p{0.6cm}p{0.6cm}p{0.75cm}|p{0.6cm}p{0.6cm}p{0.75cm}|p{0.6cm}p{0.6cm}p{0.75cm}}
    \hline
        \multirow{3}{*}{Method} & \multicolumn{6}{c|}{wiki} & \multicolumn{6}{c}{cc} \\
        \hhline{~------------}
        & \multicolumn{3}{c|}{En-Si} & \multicolumn{3}{c|}{Si-En} & \multicolumn{3}{c|}{En-Si} & \multicolumn{3}{c}{Si-En} \\
        \hhline{~------------}
         & P@1 & P@5 & P@10 & P@1 & P@5 & P@10 & P@1 & P@5 & P@10 & P@1 & P@5 & P@10 \\  \hline
        Procrustes + NN & 11.4 & 26.4 & 33.2 & 12.5 & 29.6 & 37.1 & 16.4 & 35.7 & 43.6 & 21.3 & 39.9 & 47.4 \\ 
        Procrustes + CSLS & 14.8 & 31.5 & 39.8 & 14.4 & 27.6 & 33.8 & 20.4 & 39.9 & \textbf{49.1} & 18.0 & 31.9 & 37.4 \\ \hline
        Procrustes+ refine + NN & 13.7 & 25.5 & 31.3 & 15.8 & 33.0 & 39.3 & 19.3 & 34.9 & 42.3 & \textbf{28.9} & \textbf{45.7} & 51.3 \\ 
        Procrustes+ refine + CSLS & 16.1 & 29.0 & 35.7 & \textbf{16.9} & 31.0 & 36.7 & 20.9 & 38.6 & 46.3 & 21.7 & 36.6 & 41.6 \\ \hline
        RCSLS + spectral + NN & 14.8 & 29.7 & 36.8 & 13.3 & 33.7 & 42.8 & 21.4 & 40.2 & 48.5 & 23.3 & 44.8 & 52.7 \\ 
        RCSLS + spectral + CSLS & 17.1 & 33.1 & 41.0 & 15.1 & 29.4 & 35.1 & 21.5 & 41.7 & 49.1 & 19.2 & 34.9 & 41.8 \\ \hline
        RCSLS + NN & 15.3 & 30.4 & 37.5 & 13.2 & \textbf{34.1} & \textbf{43.3} & 21.5 & 40.9 & 48.3 & 23.3 & 44.9 & \textbf{53.2} \\ 
        RCSLS + CSLS & \textbf{17.5} & \textbf{33.4} & \textbf{41.3} & 15.5 & 29.3 & 35.9 & \textbf{22.6} & \textbf{42.3} & \textbf{49.1} & 19.4 & 35.4 & 42.1 \\ \hline
    \end{tabular}
    }
    \caption{\textbf{English-Sinhala word translation average precisions} (@1, @5, @10) from 1.5k source word queries using 200k target words in \textbf{wiki} and \textbf{cc} Fasttext embeddings. \emph{Refine} is the refinement step of~\citet{conneau2017word} and, \emph{Spectral} is the \emph{Convex relaxation} step explained in \citet{joulin2018loss}. For supervised alignments, two different train-test dataset pairs have been used. }
    \label{precision_results_wiki_cc}
\end{table*}

\begin{table*}[!ht]
    \centering
    \resizebox{2\columnwidth}{!}{%
    \begin{tabular}{l|rr|rr|rr|rr|rr|rr}
    \hline
        \multirow{2}{*}{Method} & \multicolumn{10}{c|}{\citet{joulin-etal-2018-loss}} & \multicolumn{2}{c}{Ours} \\
        \hhline{~------------}
        & en-es & es-en & en-fr & fr-en & en-de & de-en & en-ru & ru-en & en-zh & zh-en & en-si & si-en \\
        \hline
        Adv.+refine & 81.7 & 83.3 & 82.3 & 82.1 & 74.0 & 72.2 & 44.0 & 59.1 & 32.5 & 31.4 & - & - \\
        Wass. Proc.+refine & 82.8 & 84.1 & 82.6 & 82.9 & 75.4 & 73.3 & 43.7 & 59.1 & - & - & - & - \\ \hline
        Procrustes & 81.4 & 82.9 & 81.1 & 82.4 & 73.5 & 72.4 & 51.7 & 63.7 & 42.7 & 36.7 & 20.4 & 18.0 \\
        Procrustes+ refine & 82.4 & 83.9 & 82.3 & 83.2 & 75.3 & 73.2 & 50.1 & 63.5 & 40.3 & 35.5 & 20.9 & \textbf{21.7} \\ \hline
        RCSLS + spectral & 83.5 & 85.7 & 82.3 & \textbf{84.1} & 78.2 & 75.8 & 56.1 & 66.5 & 44.9 & 45.7 & 21.5 & 19.2 \\
        RCSLS & \textbf{84.1} & \textbf{86.3} & \textbf{83.3} & \textbf{84.1} & \textbf{79.1} & \textbf{76.3} & \textbf{57.9} & \textbf{67.2} & \textbf{45.9} & \textbf{46.4} & \textbf{22.6} & 19.4 \\ \hline
    \end{tabular}
    }
    \caption{Extended Comparison among different alignment techniques using CSLS retrieval. 
    Here only the top-1 precision scores have been included}
    \label{alignment_methods_comparison}
\end{table*}

\begin{table*}[!ht]
    \centering
    \begin{tabularx}{\textwidth}{l|YYY}%
    \hline

        \multirow{2}{*}{Dataset} & \multicolumn{3}{c}{Scores} \\
        \hhline{~---}
        &  @1 & @5 & @10 \\
        \hline
        
        \citet{smith2016offline}: On their original eval dataset\textsuperscript{*} & 22 & 40 & 45 \\ \hline
        
        \citet{smith2016offline}+NN: On our eval dataset\textsuperscript{\textdagger} & 25 & \textbf{44} & 50 \\ \hline
        
        \citet{smith2016offline}+CSLS: On our eval dataset\textsuperscript{\textdagger} & \textbf{26} & 43 & 49 \\ \hline

        our work best results & 20 & 42 & \textbf{51} \\ \hline
        
    \end{tabularx}
    \caption{Si$\rightarrow$En Embedding Alignment Results with previous  alignment work\newline 
    \textsuperscript{*} From \citet{smith2016offline} official repository
    \textsuperscript{\textdagger} Aligned using alignment matrix given in \citet{smith2016offline} official repository and evaluated using our evaluation set. The scores can be overestimated since we do not know the exact alignment dataset used by the authors. If there is an intersection between the alignment dataset and our evaluation dataset, the scores may not represent the exact alignment accuracy.
    }
    \label{alignment_results_other_si_work}
\end{table*}

\subsection{Impact of Alignment Dataset Size}

In this section, we experimented with how the alignment dataset affects the alignment. We have experimented with an extended alignment dataset and evaluated it with the same test sets used in Section~\ref{alignment_results}. The results are reported in Table~\ref{alignment_results_dataset_size}.

\begin{table*}[!ht]
    \centering
    
    \begin{tabularx}{\textwidth}{l|c|YYY|YYY}%
    \hline
        \multirow{3}{*}{Dataset} & \multirow{3}{*}{ \makecell{Unique Src\\within 200k}} & \multicolumn{6}{c}{Retrieval} \\
        \hhline{~~------}
        &  & \multicolumn{3}{c|}{NN} & \multicolumn{3}{c}{CSLS} \\
        \hhline{~~------}
        &  & @1 & @5 & @10 & @1 & @5 & @10 \\
        \hline
        
        En-Si-para-wiki-5k & 5000  & 11.4 & 26.4 & 33.2 & 14.8 & 31.5 & 39.8 \\ \hline
        En-Si-para-wiki-full & 27846 & \textbf{17.0} & \textbf{36.1} & \textbf{45.1} & \textbf{20.2} & \textbf{42.4} & \textbf{50.9} \\ \hline
        En-Si-para-cc-5k & 5000  & 16.4 & 35.7 & 43.6 & 20.4 & 39.9 & 49.1 \\ \hline
        En-Si-para-cc-full & 27856 & \textbf{17.4} & \textbf{37.9} & \textbf{45.5} & \textbf{20.9} & \textbf{42.4} & \textbf{50.8} \\ \hline
    \end{tabularx}
    \caption{En$\rightarrow$Si Procrustes Embedding Alignment Results with different dataset sizes}
    \label{alignment_results_dataset_size}
\end{table*}

\section{Discussion and Future Work}
According to Table~\ref{precision_results_wiki_cc} and~\ref{alignment_methods_comparison}, we observe that Si-En alignment results are not on par with the high-resource language pairs. We have identified several possible reasons for this score difference.

\subsection{Impact of the embedding model size}
We observe cc Fasttext models have better alignment than wiki models. According to Table~\ref{precision_results_wiki_cc} results we can see 22.6\% @1 reduction (22.6$\rightarrow$17.5) in En-Si direction and 41.5\% @1 reduction (28.9$\rightarrow$16.9) in En-Si direction. This effect can be expected due to the comparatively low (9.7\% of cc vocabulary) vocabulary size of the Sinhala wiki FastText model (wiki-79k, cc-808k) and therefore missing a great portion of information on the Si side.

\subsection{Quality of the alignment dataset} \label{discuss_quality_of_datasets}

We have experimented only with the supervised alignment techniques in this paper and, the final alignment output solely depends on the quality of the alignment datasets that are used. Our main alignment experiments have been carried out using alignment datasets created using the base datasets provided by \citet{wickramasinghe2023sinhala} and, according to their paper, it is mentioned that the so-called~\emph{look-up score} of the datasets are not higher as expected. That indicates that there is an issue with the quality/coverage of the base dataset we used.
According to \citet{smith2016offline} the more common word pairs in the alignment dataset the better the alignment output we achieve. How we created our alignment dataset was using the English column of the En-Es MUSE~\cite{conneau2017word} alignment datasets and, therefore even if the frequent English words are included, no frequent word selection criterion was imposed on the Sinhala word selection. We assumed that by selecting the most frequent English words would indirectly lead to the most common Sinhala words. Also assumed that MUSE datasets have been created considering the most frequent words in the vocabularies~\cite{conneau2017word}.

\subsection{Alignment Techniques}
Where we do not find a proper alignment dataset, we can go for semi-supervised or unsupervised alignment techniques. The unsupervised techniques by \citet{conneau2017word} and \citet{grave2019unsupervised} have shown competitive results with the supervised techniques. Therefore our next immediate focus will be on semi-supervised and unsupervised alignment techniques.

\section{Conclusion}

The alignment dataset we used (En-Si-para-cc) has been constructed using the most frequent words in both languages as discussed in Section~\ref{discuss_quality_of_datasets}. We observed that when we do the alignment using infrequent words (i.e. alignment dictionary created without specifically considering frequent terms) the precision is worse. That is because the most frequent words' embeddings can be assumed well positioned in the embedding spaces rather than infrequent words. That observation has been reported by \citet{smith2016offline} as well.

The obtained results show that Si$\rightarrow$En alignment is better than EN$\rightarrow$Si alignment. We can explain that observation as follows. FatText English embedding space (wiki-256k, cc-2M) is considerably larger than the Sinhala embedding space (wiki-79k, cc-808k). Therefore aligning a larger embedding space onto a smaller space is lossy than the other way around given the probability of a given candidate word from the source not existing in the target is high. Further, given that Sinhala is a highly inflected language compared to English~\cite{de2019survey}, multiple morphological forms which exist in Sinhala, would invariably map to the parallel of the root word in English. Thus extenuating the viable pool of the Sinhala vocabulary to be matched to their English counterparts. We can assume that these are the reasons contributing to the drop in the resultant improvement of the @5 and @10 precision in En$\rightarrow$Si direction during the refinement procedure.

When it comes to the retrieval criterion, the CSLS gives better results than NN in most cases. Then, as far as the training objective is considered, RCSLS with CSLS as the retriever criterion has shown the best precision in most cases. This is because the core idea of RCSL alignment is to make both the training and retrieval consistent rather than using two different criteria \cite{joulin-etal-2018-loss}. 
According to \citet{aboagye-etal-2022-quantized}, the \emph{RCSL} approach by \citet{joulin-etal-2018-loss} has the highest average alignment quality/accuracy among available cross-lingual embedding alignment techniques and, from our experiments for En-Si alignment, we could verify that fact.
We have used alignment datasets with 5k unique source words for the experiments since most of the other work has been carried out with that configuration~\cite{joulin-etal-2018-loss} but, from Table~\ref{alignment_results_dataset_size} results we see that we can achieve better results by having a larger dataset.


\bibliography{anthology,custom}
\bibliographystyle{acl_natbib}

\appendix

\section{Impact of Retrieval Criterion}
\label{appendix-retrieval-methods}

\begin{table*}[!ht] 
    \centering
    \resizebox{2\columnwidth}{!}{%
    \begin{tabular}{l|rr|rr|rr|rr|rr|rr}
    \hline
       \multirow{2}{*}{Method} & \multicolumn{10}{c|}{\citet{joulin-etal-2018-loss}} & \multicolumn{2}{c}{Ours} \\
        \hhline{~------------}
        & en-es & es-en & en-fr & fr-en & en-de & de-en & en-ru & ru-en & en-zh & zh-en & en-si & si-en \\
        \hline
        Adv.+refine+NN & 79.1 & 78.1 & 78.1 & 78.2 & 71.3 & 69.6 & 37.3 & 45.3 & 30.9 & 21.9 & - & - \\
        Adv.+refine+CSLS & 81.7 & 83.3 & 82.3 & 82.1 & 74.0 & 72.2 & 44.0 & 59.1 & 32.5 & 31.4 & - & - \\ \hline
        Procrustes+NN & 77.4 & 77.3 & 74.9 & 76.1 & 68.4 & 67.7 & 47.0 & 58.2 & 40.6 & 30.2 & 16.4 & 21.3 \\
        Procrustes+CSLS & 81.4 & 82.9 & 81.1 & 82.4 & 73.5 & 72.4 & 51.7 & 63.7 & 42.7 & 36.7 & 20.4 & 18.0 \\ \hline
        RCSLS+NN & 81.1 & 84.9 & 80.5 & 80.5 & 75.0 & 72.3 & 55.3 & 67.1 & 43.6 & 40.1 & 21.5 & \textbf{23.3} \\
        RCSLS+CSLS & \textbf{84.1} & \textbf{86.3} & \textbf{83.3} & \textbf{84.1} & \textbf{79.1} & \textbf{76.3} & \textbf{57.9} & \textbf{67.2} & \textbf{45.9} & \textbf{46.4} & \textbf{22.6} & 19.4 \\ \hline
    \end{tabular}
    }
    \caption{Extended Comparison nearest neighbour (NN) and CSLS retrieval Criteria. 
    Here only the top-1 precision scores have been included}
    \label{retrieval_methods_comparison}
\end{table*}

Table~\ref{retrieval_methods_comparison} shows a comparison of how Si-En aligned embeddings behave with different retrieval criteria with other language pairs. In all the other language pair results given in~\citet{joulin2018loss}, the RCSLS criterion outperforms the NN criterion in both directions but, in our case, Si$\rightarrow$En direction, NN has shown the best results while En$\rightarrow$Si shows the best results with CSLS. This effect can be clearly seen in Table~\ref{precision_results_wiki_cc} as well. \citet{joulin2018loss} says, \emph{RCSLS transfers some local information encoded in the CSLS criterion to the dot product.} to establish a suggestion as to why RCSLS outperforms NN in their results but, it seems RCSLS need not be the best retriever criterion for all the cases and, could depend on the language pair and the alignment direction.




\end{document}